\documentclass[ijoc,nonblindrev]{informs3} 

\OneAndAHalfSpacedXII 

\usepackage{natbib}
 \bibpunct[, ]{(}{)}{,}{a}{}{,}
 \def\bibfont{\small}%

\usepackage{algorithmicx,algorithm}
\usepackage[noend]{algpseudocode}
\usepackage{graphicx}
\usepackage{amsmath}
\usepackage{multirow}
\usepackage{amsmath}
\usepackage{amssymb}
\usepackage{tikz}
\usepackage{amsbsy}

\usepackage{enumitem} 
\usepackage{epstopdf} 
\usepackage{graphicx}
\usepackage{footnote}
\usetikzlibrary{arrows}
\newcommand{\bl}{\boldsymbol}
  
\usepackage{subcaption}
\graphicspath{{./figures/}}

\TheoremsNumberedThrough     

\EquationsNumberedThrough    

\begin{document}


\RUNAUTHOR{Meng, Wang and Ng}

\RUNTITLE{Combined Global and Local Search for Optimization with Gaussian Process Models}

\TITLE{Combined Global and Local Search for Optimization with Gaussian Process Models}

\ARTICLEAUTHORS{%
\AUTHOR{Qun Meng}
\AFF{Department of Industrial Systems Engineering and Management, National University of Singapore, Singapore 117576, \EMAIL{sjtumengqun@gmail.com}
}
\AUTHOR{Songhao Wang}
\AFF{Department of Information Systems and Management Engineering, Southern University of Science and Technology, No. 1088 Xueyuan Ave., Shenzhen, China  518000, \EMAIL{wangsh2021@sustech.edu.cn}
}
\AUTHOR{Szu Hui Ng}
\AFF{Department of Industrial Systems Engineering and Management, National University of Singapore, Singapore 117576, \EMAIL{isensh@nus.edu.sg}
}
} 

\ABSTRACT{%
Gaussian process (GP) model based optimization is widely applied in simulation and machine learning. In general, it first estimates a GP model based on a few observations from the true response and then employs this model to guide the search, aiming to quickly locate the global optimum. Despite its successful applications, it has several limitations that may hinder its broader usage. First, building an accurate GP model can be difficult and computationally expensive, especially when the response function is multi-modal or varies significantly over the design space. Second, even with an appropriate model, the search process can be trapped in suboptimal regions before moving to the global optimum due to the excessive effort spent around the current best solution. In this work, we adopt the Additive Global and Local GP (AGLGP) model in the optimization framework. The model is rooted in the inducing-points-based GP sparse approximations and is combined with independent local models in different regions. With these properties, the AGLGP model is suitable for multi-modal responses with relatively large data sizes. Based on this AGLGP model, we propose a Combined Global and Local search for Optimization (CGLO) algorithm. It first divides the whole design space into disjoint local regions and identifies a promising region with the global model. Next, a local model in the selected region is fit to guide detailed search within this region. The algorithm then switches back to the global step when a good local solution is found. The global and local natures of CGLO enable it to enjoy the benefits of both global and local search to efficiently locate the global optimum. 
}%


\KEYWORDS{simulation optimization; Gaussian process; global and local search}

\maketitle

%

\section{Introduction}
Gaussian process model based optimization, also known as Bayesian optimization (BO), is an effective approach to optimize black-box functions. By black-box functions, we refer to objective functions that have no explicit forms and can only be observed at arbitrary inputs. We further assume that observations are noisy and expensive to evaluate. The black-box function optimization can be formalized as
\begin{equation}
\label{objective}
\text{min}_{{\bf x}\in \mathbb{X}_{\Omega}} \ \  \mathbb{E}_\xi[y({\bf x},\xi) ],
\end{equation}
where ${\bf x}\in \mathbb{X}_{\Omega} \subset \mathbb{R}^d$ ($\mathbb{X}_{\Omega}$ is the design space assumed compact) is the design choice and $\xi$ represents the randomness, which is a noise following normal distribution whose variance depends on ${\bf x}$. BO has been successfully applied in many problems. These include inventory problems where the decision-maker uses a simulation model to determine the best inventory level (the decision variable ${\bf x}$) to maintain, in order to maximize the expected profit (the response $y$). It has also been widely applied to tune hyperparameters for algorithms like neural networks \citep{snoek2012practical}, where the response $y$ in \eqref{objective} typically represents the training error with respect to a set of hyperparameters ${\bf x}$. Other popular application areas of BO include sensor networks, robotics, advertising, recommendation and reinforcement learning (see \cite{shahriari2016taking} for a survey).

BO utilizes a GP model (also known as a kriging model) that is built with observations at sequentially selected design points. This model is essentially a statistical approximation of the baseline response and is considered as a type of metamodel. Metamodels summarize available information to predict the performance of unobserved points. Therefore, they are often used to guide optimizations for expensive responses. Compared with other metamodels, the GP model is very flexible with fewer assumptions required on the response structure and `more resistant to overfitting than general interpolation methods' (such as the neural network) \citep{ankenman2010stochastic}. Furthermore, the GP model provides an analytical form of its prediction uncertainty which can be used to determine the accuracy of the prediction and importantly, also help determine new design points to improve the model fitting. Typically, in BO, the next design point is chosen to maximize an acquisition function, which measures the utility of candidate points, computed with the GP model. After selection, the new observation is obtained and the GP model is updated accordingly. 

BO can be categorized with respect to the acquisition functions adopted. Popular choices include Expected improvement (EI) \citep{jones1998efficient}, GP upper confidences bound \citep{srinivas2009gaussian}, stepwise uncertainty reduction \citep{picheny2015multiobjective}, knowledge gradient \citep{frazier2009knowledge} and entropy search \citep{hennig2012entropy} (see \cite{frazier2018tutorial,shahriari2016taking} for reviews). Among these approaches, EI is easy to perform and perhaps the most widely-used due to its ability to automatically balance between exploitation (searching around the current best solution) and exploration (searching less-explored regions). EI was originally proposed by \cite{jones1998efficient} in the efficient global optimization (EGO) algorithm to optimize deterministic functions. This acquisition function and algorithm were later adapted for stochastic simulations with homogeneous noise \citep{huang2006global}. More recently, the Two Stage Sequential Optimization (TSSO) \citep{quan2013simulation} and the extended-TSSO \citep{Pedrielli2018etsso} were proposed for the more general heteroscedastic noise case, where the noise levels at different input levels are not the same. These algorithms consist of two stages, namely the searching stage and the allocation stage, where the searching stage focuses on selecting the next design point based on the EI criterion, and the allocation stage focuses on allocating additional replications of function evaluations to existing design points to reduce the noise levels. Similar to these works, we adopt the EI criterion here as the acquisition function in our algorithm.

Despite numerous successful applications, BO has a few limitations that can hinder its broader usage. To illustrate them, we use an example of an agent-based simulation used in maritime safety. Here, real-time decisions on the trajectory for a vessel, characterized by four parameters including two turning angles and two traveling speeds, are required to minimize its probability of conflict with other vessels when sailing through a region (see detailed description in Section 6.3). In the simulated high-congestion region, a small turn of the vessel can result in a very different conflict environment, thus, resulting in a multi-modal response that can change dramatically across the design space. This raises a few difficulties for the traditional BO approach. First, to build an accurate GP model is challenging for such a non-homogeneous response, unless a large number of design points are used. This, in turn, brings a strain on the simulation resources and the computational cost of fitting the GP model as it increases cubically with the number of design points \citep{quinonero2005unifying}. For such real-time decision problems, the large computational burden can diminish the advantage of BO. Hence, fast approximations of the GP model are required in such situations. Second, even with a proper GP model, the optimization can be hard. Although EI ensures global convergence under mild conditions \citep{bull2011convergence}, it tends to over-exploit local optimal regions \citep{ranjan2011computationally}, which can result in excessive efforts expensed around the current best before opportunities are given to explore other regions where better solutions may exist. With a finite budget, this can cause the algorithm to return a sub-optimal solution, especially if the response has many local optimums like the one in our maritime safety example. 
 
To address these challenges, in this paper, we develop a new BO algorithm, the Combined Global and Local search for Optimization algorithm (CGLO), which combines an iterative global and local search processes to optimize responses with heteroscedastic noise. CGLO leverages on a recently proposed additive GP model, the AGLGP model \citep{meng2015additive} and exploits the model's local and global structure to more efficiently guide the search. In this approach, the model fitting is faster and enables more flexibility in capturing the multi-modal response surface. Furthermore, the built-in mechanism in the switching criteria of the algorithm helps to avoid the over-exploitation of any single region.

While several fast approximation methods have been proposed such as sparse approximations \citep{quinonero2005unifying} and local kriging, none is able to adequately capture both global trends and local fluctuations of many nonstationary responses. Combined models \citep{snelson2007local,ba2012composite}, on the other hand, are able to better capture these global and local trends, but they still assume similar local trends throughout the design space, thus restricting their modeling flexibility.  Different from these works, the AGLGP model explicitly models the local residual structures independently and non-identically to enable this flexibility.  

Our main contributions in this paper can be summarized as follows:  
\begin{enumerate}
	\item We provide the full development of the AGLGP model and propose a faster approach to build the model;
	\item We adapt the EI criterion with the AGLGP model to develop a new algorithm with global and local search steps;
	\item We provide the asymptotic convergence result for the proposed CGLO algorithm and test its finite-time performance with several examples.
\end{enumerate}

The rest of this paper is organized as follows.  Section 2 provides an overview of our CGLO algorithm and its key components. Next, Section 3 develops in detail the first key component, the AGLGP model. Subsequently, Section 4 describes the CGLO algorithm and Section 5 states its convergence results. Section 6 provides numerical tests of CGLO including some test functions and a maritime navigational safety application. Finally, section 7 concludes the paper.  The proofs of all lemmas and theorems and all the appendices are provided in the online supplementary material.

\section{Overview of CGLO Algorithm}

The key idea of CGLO is to leverage on the local and global structure of the AGLGP model to direct a global exploration across the space with its global component, and to build the detailed AGLGP model in local regions to search more efficiently and accurately in these regions (without the need of a detailed estimation of the full GP model). The algorithm iterates between this global and local search levels, providing a cycle of learning between them to improve the quality of the search without overburdening the cost of estimating the model, and a switching criterion is used to decide when the CGLO switches back to the global level to avoid getting trapped in a local optimal. We illustrate the key ideas of CGLO with a simple example below.

\begin{figure}[htbp]
	\vspace{-0.6cm}
	\centering 
	\includegraphics[scale=0.6]{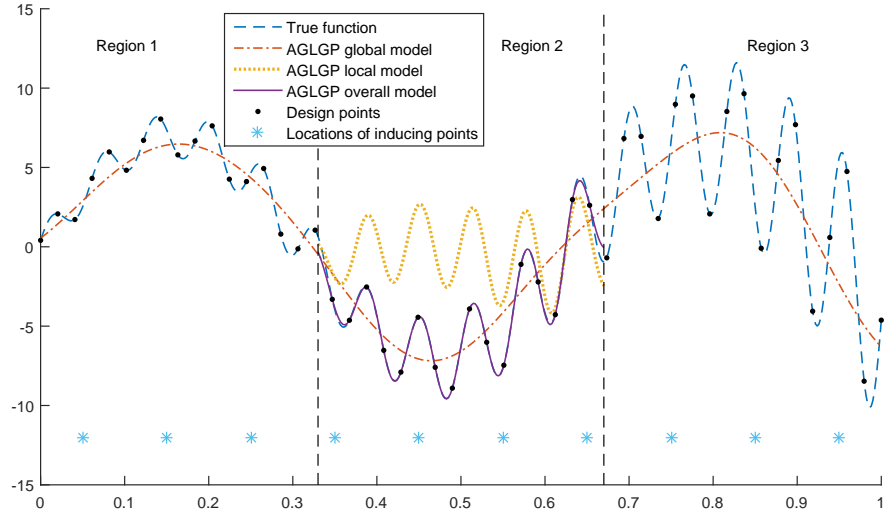}
	\caption{Illustrating example of the AGLGP model used in the CGLO algorithm}\label{ill}
\end{figure}

Consider an example where the function $y(x)=\cos(100(x-0.2))\exp(2x)+7\sin(10x)$ (see Figure \ref{ill}) is to be minimized. Here we see that the function has a global trend and also noisy local variations. If we directly use a full GP-EI algorithm, the search can be misleading at the start (with a small number of points) and potentially trapped in one of the many local optimums. Moreover, the full GP model becomes computationally expensive as the design points increase as the search continues. Here instead, we first fit the AGLGP model. This model form consists of a global term capturing the global trend and separate local models capturing the remaining process in each local region. In the global model, to capture the global picture, the information of the design points are summarized by the inducing points and local fluctuations are smoothed out (see red dash-dotted line in Figure 1). To capture the local residuals, the whole design space is divided into $K$ non-overlapping regions and local models are fitted separately to the residuals in each region (see the yellow dotted line in Figure 1 for Region 2). As the inducing points (to build the global model) and design points in each local region (to build the local model) are much fewer than the total number of design points, the computational cost to fit the model is greatly reduced. 

The CGLO algorithm then leverages these model features in two iterative steps: a global step and a local step. The global step focuses on identifying promising regions and is guided by the global model, as it is less likely to be influenced by local optimums and able to better focus on regions where the global trends are low.  With an initial model fit, we are able to identify Region 2 as a promising region. A local model in this region is then incorporated with the global model to form an overall model (see the purple line in Figure 1). This overall AGLGP model is used to guide a more refined in-depth search within this selected promising region to identify optimums within this region. A switching criterion is then used to determine the effort to expense in this region before reverting back to the global step. The idea behind this switch criterion is to allow sufficient resources in the local step to identify promising local minimums but to avoid over expense on negligible local improvements. This switch thus helps CGLO jump out of the current exploiting regions to reconsider the other candidate regions where better solutions may exist and avoid being trapped in one sub-optimal region. After each local step, the global model is updated with the new observations to improve the directed search in the global step. The proposed procedure then alternates between the global and local searches to more efficiently explore and exploit the entire region.  

From this example, we see that there are two main components to CGLO: 
\begin{enumerate}
	\item A fast AGLGP model that jointly models the global trend and the local fluctuations.
	\item An algorithm that leverages on this model to develop global and local search steps.
\end{enumerate}
In the following two sections, we will develop in detail these two components. 

\section{AGLGP Model Development}
The AGLGP model has been shown to work well for multi-modal responses and those that vary significantly across the design space \citep{meng2015additive}. In this section, we first review the basics of the AGLGP model in Section 3.1. In Section 3.2, we describe a fast estimation approach for the AGLGP model that is more suitable for CGLO. It reduces the complexity of traditional BO from $O(n^3)$ to $O(nm^2+nB^2)$, where $n$, $m$ and $B$ are the number of design points, the number of inducing points and the number of design points in each local region, respectively \citep{meng2015additive}. 

\subsection{AGLGP Model Review}
The basic idea of AGLGP model can be seen from Figure 1. It models the response with a global term (the red dash-dotted line) and independent local terms (the yellow dotted line), which captures the local residual process. Built by the inducing points (the blue stars in Figure 1), the global model has long lengthscale to capture the global trend while the separate local models are developed with shorter lengthscales. The following parameters in Table \ref{AGLGPpra} are used to develop the AGLGP model.
\begin{table}[h]\small
	\centering
	\caption{Notations for AGLGP model}\label{AGLGPpra}
	\begin{tabular}{c|c}
		\hline
		Notation & Definition\\
		\hline
		$\mathbf{x_g}$; $\mathbf{y_g}$ & The $m$ global inducing points $(\mathbf{x_g} ^1,\cdots,\mathbf{x_g}^m)$; The latent global 'observations' at $\mathbf{x_g}$\\
		$\mathbf{D}_k$ & The $k$th local region $k=1,\cdots,K$ where $\mathbf{D}_j\bigcap \mathbf{D}_k=\emptyset$ for $ k\neq j$ and $\cup_{k=1}^K\mathbf{D}_k=\mathbb{X}_{\Omega}$ \\
		$\mathbf{x}^k$; $\mathbf{y}^k$ & The $N(\mathbf{D}_k)$ design points in region $\mathbf{D}_k$ $(\mathbf{x}^1_l,...,\mathbf{x}^{N(\mathbf{D}_k)}_l)$; The response observations at $\mathbf{x}^k$	\\
		$n$ & The total number of design points: $n=N(\mathbf{D}_1)+...+N(\mathbf{D}_k)$\\
		$\mathbf{X}$; $\mathbf{Y}$  & The vector of all design points $(\mathbf{x}^1,...,\mathbf{x}^K)$; The response observations at $\mathbf{X}$\\
		
		\hline
	\end{tabular}
\end{table}
The AGLGP model assumes the stochastic simulation response at $\mathbf{x}$ can be represented as:
\begin{equation}
y(\mathbf{x})=f(\mathbf{x})+\epsilon(\mathbf{x})=f_{global}(\mathbf{x})+\sum_{k=1}^Kw_k(\mathbf{x})f^k_{local}(\mathbf{x})+\epsilon(\mathbf{x}),\quad
w_k(\mathbf{x})=\left\{\begin{array}{rcl} 
1,\mathbf{x}\in \mathbf{D_k} \\ 
0,\mathbf{x}\notin \mathbf{D_k}
\end{array}\right.\label{model}
\end{equation}
Here, the process mean of the response $f(\mathbf{x})$ consists of a global trend process $f_{global}(\mathbf{x})$ and $K$ local residual processes $f^k_{local}(\mathbf{x})$ in region $\mathbf{D}_k$. These processes are assumed to be piecewise independent. $f_{global}(\mathbf{x})$ is assumed to be a Fully Independent Training Conditional (FITC) GP approximation \citep{quinonero2005unifying} with mean $\mu$ and covariance function $R_g(\cdot)$, while $f^k_{local}(\mathbf{x})$ is assumed to be a stationary GP in local region $\mathbf{D}_k$ with mean 0 and covariance function $R_l^k(\cdot)$, where
\begin{gather}
R_g(\mathbf{x}_i,\mathbf{x}_j)=\sigma^{2}corr_g(\mathbf{x}_i,\mathbf{x}_j,\pmb{\theta}),\quad
R_l^k(\mathbf{x}_i,\mathbf{x}_j)=\tau_k^{2}corr^k_l(\mathbf{x}_i,\mathbf{x}_j,\bl{\alpha}_k).\notag
\end{gather}
$\sigma^{2}$ and $\tau_k^{2}$ are the GP model variances of $f_{global}(\mathbf{x})$ and $f^k_{local}(\mathbf{x})$, respectively; $corr_g(\cdot)$ and $corr^k_l(\cdot)$ are the correlation function of $f_{global}(\mathbf{x})$ and $f^k_{local}(\mathbf{x})$, respectively. A widely applied correlation function is the Gaussian correlation: $corr(\mathbf{x}_i,\mathbf{x}_j,\pmb{\theta})=\exp[\sum_{k=1}^{d}-\theta_i (\mathbf{x}_{i,k}-\mathbf{x}_{j,k})^2 ]$, where $\mathbf{x}_{i,k}$ is the $k$-th entry of $\mathbf{x}_i$. To sufficiently consider the nonstationarity of the function, in different local models, the covariance structure can be different. Furthremore, in the AGLGP model, and additional constraint is added: $\mathbf{0}\leq\bl{\theta}\leq\bl{\alpha}$, i.e., the global model has longer lengthscale parameter than the local models. This will ensure a smoother global model for the global trend. In addition, $\epsilon(\mathbf{x})$ is a normal noise $\mathcal{N}(0,\sigma_\epsilon^2(\mathbf{x}))$ independent from $f_{global}(\mathbf{x})$ and $f^k_{local}(\mathbf{x})$.  

As a latent process, $f_{global}(\mathbf{x})$ only depicts a general and rough global picture of the true function, and it is assumed to be a deterministic model. Suppose $\mathbf{x_g}$ and their global values $\mathbf{y_g}$ are known. The global predictor at any given point $\mathbf{x}_0$ can then be written as
\begin{equation}
\widehat{y}_{global}(\mathbf{x}_0)=\mu+\mathbf{g}(\mathbf{x}_0)\mathbf{'G}_m^{-1}(\mathbf{y_g}-\mathbf{1}'\mu),\label{gm}
\end{equation} 
where $\mathbf{g}(\mathbf{x}_0)=(R_g(\mathbf{x}_0,\mathbf{x_g}^1),\cdots, R_g(\mathbf{x}_0,\mathbf{x_g}^m))$, $\mathbf{G}_m$ is covariance matrix at $\mathbf{x_g}$ with $(i,j)th$ entry $R_g(\mathbf{x_g}^i,\mathbf{x_g}^j)$.
We thus have the global predictor at all design points: $\mathbf{\widehat{y}}_{global}=(\widehat{y}_{global}(\mathbf{x}_1),...,\widehat{y}_{global}(\mathbf{x}_n))$. The residuals $\mathbf{y}_{local}=\mathbf{Y}-\mathbf{\widehat{y}}_{global}$ is then modeled by another stochastic GP model $y_{local}(\mathbf{x})=\sum_{k=1}^Kw_k(\mathbf{x})f^k_{local}(\mathbf{x})+\epsilon(\mathbf{x})$. This local model captures the local biases of the global model in each local region and the inherent stochastic noise in the system. As we focus on stochastic simulations, $\mathbf{Y}$ is the sample mean of the replications taken at each design point. Given $K$ local regions $\mathbf{D}_1,\cdots,\mathbf{D}_K$, the local predictor at any given point $\mathbf{x}_0$ is given by
\begin{equation}
\widehat{y}_{local}(\mathbf{x}_0)=\mathbf{l}_k' (\mathbf{x}_0)\mathbf{(L}_k+\mathbf{\pmb{\Sigma}}_{\epsilon,k})^{-1}\mathbf{y}^k_{local}, \forall \,\mathbf{x}_0\in \mathbf{D}_k,\label{lm}
\end{equation}
where $\mathbf{l}_k(\mathbf{x}_0)=(R_l^k(\mathbf{x}_0,\mathbf{x}^1_l),\cdots, R_l^k(\mathbf{x}_0,\mathbf{x}^{N(\mathbf{D}_k)}_l))$ and $\mathbf{L}_k$ is the covariance matrix within a local region $\mathbf{D}_k$ whose $(jh)th$ element $R_l^k(\mathbf{x}_l^h,\mathbf{x}^j_l), \forall \mathbf{x}^j_l,\mathbf{x}^h_l \in \mathbf{x}^k$. $\bl{\Sigma}_{\epsilon,k}=diag({\sigma}^2_{\epsilon}(\mathbf{x}^1_l),...,{\sigma}^2_{\epsilon}(\mathbf{x}^{N(\mathbf{D}_k)}_l))$. $\mathbf{y}^k_{local}$ is the local residuals in region $\mathbf{D}_k$ at points $\mathbf{x}^k$.

From (\ref{gm}) and (\ref{lm}), the overall AGLGP predictor can be written as $\widehat{y}_{global}(\mathbf{x}_0)+\widehat{y}_{local}(\mathbf{x}_0)$. This predictor, however, can only be obtained when the latent global evaluations $\mathbf{y_g}$ is known, which is impractical since typically only the overall response observations $\mathbf{Y}$ are available. The overall AGLGP predictor and predictive variance for $\mathbf{x}_0\in \mathbf{D}_k$ is obtained by integrating out $\mathbf{y_g}$ as (we write $\mathbf{g}(\mathbf{x}_0)$ and $\mathbf{l}_k (\mathbf{x}_0)$ as $\mathbf{g}$ and $\mathbf{l}_k$ for short):
\begin{equation}
\label{AGLGPpre}
\widehat{y}(\mathbf{x}_0)=\mu+(\mathbf{g}\mathbf{'G}_m^{-1}\mathbf{G}_{mn}+ \mathbf{l'} )[\mathbf{G}_{nm}\mathbf{G}_m^{-1}\mathbf{G}_{mn}+\bl{\Lambda} +\mathbf{L}_n +\bl{\Sigma}_{\epsilon}]^{-1}(\mathbf{Y}-\bl{\mu}),
\end{equation}
\begin{equation}
\label{AGLGPvar}
\widehat{s}^2(\mathbf{x}_0)=\sigma^{2}+\tau_k^{2}-(\mathbf{g}\mathbf{'G}_m^{-1}\mathbf{G}_{mn}+ \mathbf{l'} )[\mathbf{G}_{nm}\mathbf{G}_m^{-1}\mathbf{G}_{mn}+\bl{\Lambda} +\mathbf{L}_n +\bl{\Sigma}_{\epsilon}]^{-1}(\mathbf{g}'\mathbf{G}_m^{-1}\mathbf{G}_{mn}+ \mathbf{l}')'.
\end{equation}
Denote the $j$th design point in $\mathbf{X}$ as $\mathbf{x}_j$. Here, $\mathbf{G}_{mn}$ is $m\times n$ covariance matrix with $(i,j)th$ element $R_g(\mathbf{x_g}^i,\mathbf{x}_j)$, $\mathbf{G}_n$ is $n\times n$ covariance matrix with $(i,j)th$ element $R_g(\mathbf{x}_i,\mathbf{x}_j)$, $\bl{\Lambda}=diag\{\mathbf{G}_n-\mathbf{G}_{nm}\mathbf{G}_m^{-1}\mathbf{G}_{mn}\}$, $\mathbf{l}'=(\mathbf{0}_{{N(\mathbf{D}_1)}},...,\mathbf{0}_{{N(\mathbf{D}_{k-1})}},\mathbf{l}_k',...,\mathbf{0}_{{N(\mathbf{D}_K)}})$ ($\mathbf{0}_{{N(\mathbf{D}_i)}}$ is a $1\times {N(\mathbf{D}_i)}$ vector of all 0, $\mathbf{L}=diag\{\mathbf{L}_1,\cdots, \mathbf{L}_K\}$ and $\bl{\Sigma}_{\epsilon}=diag({\sigma}^2_{\epsilon}(\mathbf{x}^1),...,{\sigma}^2_{\epsilon}(\mathbf{x}^{n}))$.

In order to develop the AGLGP model \eqref{AGLGPpre}, two important steps have to be taken. First, the design space has to be divided into local regions for the local models (see the boundaries between the regions in Figure 1). Second, a smaller set of inducing points (the blue stars in Figure 1) has to be determined from the large set of design points to fit the global model. To divide the design space, \cite{meng2015additive} suggested to first divide the existing design points into $K$ groups with the $k$-means clustering technique and then build the separation boundaries between each pair of groups with the support vector machine (SVM) method. The regions enclosed by these boundaries are treated as local regions. Such a scheme helps to better approximate the assumption of independence of the local processes across regions as SVM tries to provide the largest separation between the grouped design points. In practice, to save the computing effort expensed for SVM (which is about $O(n^3)$, the same as fitting a full GP model), the boundary between two clusters can also be chosen as the bisecting hyperplane of the line segment whose ends are the two clusters' centers and this is used in our numerical test. The selection of inducing points is essentially to further divide the design points in the same local region to smaller clusters, such that the points in the same cluster are `similar' and can be represented by a single inducing point. The simplest way to do this is to adopt the traditional clustering techniques, like the $k$-means. These techniques, however, are mostly unsupervised learning methods in that they don't consider the response values (or labels). For the AGLGP model, the response values are available at the design points and considering these values when deciding the inducing points can better summarize the points in the same cluster. To achieve this, the original AGLGP work \citep{meng2015additive} proposed a new method to first group the points in each local region by their responses, i.e., the points in each group have close $y$ values. Then, the points in each group are further clustered by their $x$ values. The details of the methods to partition the space and select the inducing points proposed in the original AGLGP work are provided in Appendix A. We highlight here that although these suggested methods are used here, the CGLO algorithm is not restricted to these methods. Various other techniques for 
clustering, like the model-based clustering scheme \citep{fraley2002model,peng2018efficient}, partitioning, such as random division or partitioning with Mondrian process \citep{roy2008mondrian}, and inducing points selection, such as those used in the sparse GP model approximations \citep{quinonero2005unifying}, can also be applied. 

Before closing this section, we highlight that although the overall AGLGP predictor \eqref{AGLGPpre} provides an overall prediction, it does not provide separate predictions for the global trend and the local residues. As we shall see in the following sections, it is desirable to develop separate predictions for these two, like \eqref{gm} and \eqref{lm}, to integrate them into CGLO so that separate search processes in the global and local levels can be conducted. This motivates us to further develop a two-stage AGLGP model in Section 3.2.

\subsection{Development of a Fast Two-Stage Model Fitting Procedure for the AGLGP Model}
Although the AGLGP model is developed upon a global and local model structure, estimating the overall likelihood provides only an overall prediction (see \eqref{AGLGPpre}). This is useful for model inferences and predictions. However, to leverage the individual components to facilitate the global and local searches, separate predictions for the two different levels are required. Here, we develop a two-stage AGLGP model fitting procedure which separates the predictions and is computationally faster.

The two-stage approach will first construct the global model $\hat{y}_g(\mathbf{x}_0)$ after integrating out only the latent variable $\mathbf{y_g}$. The predicted residuals $\mathbf{y}_{local}=\mathbf{Y}-\hat{\mathbf{y}}_{global}$ for the local model are then obtained. The parameters of the local model are then obtained from the likelihood of the local model with the predicted residuals. The two stage predictive mean $\hat{y}(\mathbf{x}_0)=\hat{y}_g(\mathbf{x}_0)+\hat{y}_l(\mathbf{x}_0)$ and predictive variance for $\mathbf{x}_0\in \mathbf{D}_k$ can be derived as
\begin{gather}
\hat{y}_g(\mathbf{x}_0)=\mu+\mathbf{g}'\mathbf{Q}_m^{-1}\mathbf{G}_{mn}(\boldsymbol{\Lambda+\Sigma}_{\epsilon})^{-1}(\mathbf{Y-1'}\mu),\label{gpred}\\
\hat{s}_g^2(\mathbf{x}_0)=\sigma^{2}-\mathbf{g}'\mathbf{G}_m^{-1}\mathbf{g}+\mathbf{g}'\mathbf{Q}_m^{-1}\mathbf{g} \label{gmse}\\
\hat{y}_l(\mathbf{x}_0)=\mathbf{l}'(\mathbf{L}+\boldsymbol{\Sigma}_{\epsilon})^{-1}(\boldsymbol{\Lambda+\Sigma}_{\epsilon}-\mathbf{G}_{nm}\mathbf{Q}_m^{-1}\mathbf{G}_{mn})(\boldsymbol{\Lambda+\Sigma}_{\epsilon})^{-1}(\mathbf{Y-1'}\mu),\\
\hat{s}_l^2(\mathbf{x}_0)=\tau_k^{2}-\mathbf{l}'(\mathbf{L}+\bl{\Sigma}_{\epsilon})^{-1}\mathbf{l}\label{gmsee}
\end{gather}
where $\mathbf{Q}_m=\mathbf{G}_m+\mathbf{G}_{mn}(\boldsymbol{\Lambda+\Sigma}_{\epsilon})^{-1}\mathbf{G}_{nm}$. It can also be shown that the two-stage predictor is an unbiased predictor given the parameters. Details of the two-stage development and estimation are described in Appendix B. In real applications, multiple evaluations are conducted at each design points. The observation $\mathbf{Y}$ and noise variance $\sigma_{\epsilon}^2(\mathbf{x})$ in \eqref{gpred} $\sim$ \eqref{gmsee} are then replaced with the corresponding sample mean and variance.

Overall, this two-stage approach is computationally faster than the one-stage approach \citep{meng2015additive} as it optimizes the global and local parameters independently (see Appendix B). The one-stage estimation approach takes into account the interaction between the global model and the local model, and there are $(3K+1)(d+1)^2$ items in the Hessian matrix that needs to be calculated. In our proposed two-stage estimation approach, only a $(d+1)\times (d+1)$ Hessian matrix of the second-order derivative of the log-likelihood function needs to be calculated for the global model, and the local model estimation requires additional calculation of $K$ $(d+1)\times (d+1)$ Hessian matrices. Hence, in the iterative maximization of the likelihood function, the two-stage estimation will require less than half the effort to calculate its Hessian matrix in each iteration. Therefore, in practice, the one-stage approach requires a higher-dimensional likelihood optimization problem, which is more difficult to solve and computationally more expensive. In contrast, the two-stage approach proposed here is cheaper by separating the global and local parameters estimations and can guide the search in CGLO with separate global and local models more efficiently.

\section{Combined Global and Local Search for Optimization}
This section introduces the CGLO algorithm built on the global and local features of the AGLGP model in each iteration. CGLO is an iterative algorithm, where each iteration is comprised of two search steps and an Allocation Step. Among the two search steps, the Global Search Step finds promising regions with the global model while the Local Search Step exploits within these promising regions to find the optimal point with the overall model. The Allocation Step then focuses on driving down noise levels to improve the estimations in both steps. Further, a switching criterion is used to decide when to switch back to the Global Search Step from the Local Search Step and advance iterations to avoid over-exploitation within each iteration. An overview of CGLO is outlined in Algorithm 1 (a flow chart of CGLO is provided in Appendix D).

\textbf{Algorithm 1} (CGLO)
\begin{itemize}
	\item[] \textit{Step 1:} \textit{(Initialization)} Fit an initial AGLGP model with space-filling design strategy and use cross-validation technique to ensure it is valid.
	\item[] \textbf{Repeat Steps 2$\sim$3 until the budget is exhausted.}
	\item[] \textit{Step 2.1:} \textit{(Global Search Step)} Select a point $\mathbf{x}^g_0$ among $\mathbb{X}_{G}$, a set of candidate points for the global search, with the global model searching criterion. Identify the current promising region $\mathbf{D}_{k,t}$, where $\mathbf{x}^g_0\in \mathbf{D}_{k,t}$.
	\item[] \textit{Step 2.2:} \textit{(Local Search Step)} Search for the next evaluation point with a local search criterion based on the overall model fitted in $\mathbf{D}_{k,t}$, and then evaluate this point with $r_{min}$ replications. This step is repeated until the switching criterion is satisfied. 	
	\item[] \textit{Step 3:} \textit{(Allocation Step)} Allocate additional evaluations to the selected design points. 	 
	\item[] \textit{Step 4:} \textit{(Return Optimal Solution)} Report the best point (with the optimal sample mean $\bar{y}^*$).
\end{itemize}
\vspace{0.4cm}

The key notations for CGLO are listed in Table 2 (the steps where they appear are provided in the brackets). They are classified into two groups. The first group includes the `inputs' determined by the users to run the algorithm, and the second group includes the `parameters' that will be dynamically determined by CGLO. 
\begin{table}[h]\small
	\centering
	\caption{Key notations for CGLO}\label{CGLOpra}
	\begin{tabular}{c|c}
		\hline
		Inputs & Definition\\
		\hline
		$K$ & the number of local regions\\ 
		$T$ & the total budget (total number of affordable simulation replications)\\ $\mathbb{X}_{G}$ & The candidate points set for the global search (Step 2.1)\\$v$ & the parameter controlling the penalty in the gEI function (Step 2.1)\\ $r_{min}$ & the number of replications allocated for a newly selected point (Step 2.2)\\ $B_{t,a}^2$ & the budget consumed to improve local estimate in iteration $t$ (Step 3)\\ \hline Parameters & Definition \\ \hline 
		$\mathbf{x}_g^0$ & the best point in the Global Search Step	(Step 2.1)\\ 		
		$n_a(\mathbf{x})$ &  the number of neighbor design points of $\mathbf{x}$ (Step 2.1)\\ $\mathbf{D}_{k,t}$  & the most promising region in iteration $t$ (Step 2.1)\\ $n_t$ & the number of design points selected in iteration $t$ (Step 2.2)\\ $N_t$ & the total number of design points selected in the first $t$ iterations (Step 3)\\ $N_t(\mathbf{D}_{k,t})$ & the total number of design points in $\mathbf{D_{k,t}}$ (Step 3)\\ $M_t(\mathbf{x})$ & the total number of replications received by $\mathbf{x}$ in the first $t$ iterations (Step 3)\\ $N_{OCBA}$ & the number of replications added by the OCBA technique in the Allocation Step (Step 3)\\ $\bar{y}_i$ & the sample mean of $\mathbf{x}_i$ (Step 3)\\ $\mathbf{x}_b$ & the point with the lowest sample mean in the current promising region (Step 3)\\ $\bar{y}_b$ & the sample mean of $\mathbf{x}_b$ (Step 3)\\	$B_{t,a}^1$ & The budget consumed to improve global estimate in iteration $t$ (Step 3)\\	
		\hline
	\end{tabular}
\end{table}
We first discuss how to decide two key inputs, $K$ and $r_{min}$. The remaining inputs will be discussed later when we introduce Steps 2 and 3 in detail in Sections 4.1 to 4.3. 

\textbf{$K$ and local regions}: It is worthwhile to note that in Algorithm 1, the local regions are decided based on the initial design points and are kept fixed over the iterations. This removes further computations that would be required in the redivision of regions after each iteration. Moreover, in the fitting of a good reliable GP model, a reasonable number of design points is required, and thus, continued division of the regions can lead to smaller and smaller numbers of points in some local regions, which results in less reliable local models. This will be more severe if the budget is constrained and the total number of affordable design points is limited, which is common in situations where BO has been applied and proved to be competitive. We, therefore, consider fixed local regions in this work to solve our motivating navigational safety problem where the decision is required within a very short period of time. Nonetheless, when the budget is large, CGLO can be modified to update the local regions after Step 3 in each iteration. We provide a modified approach to accommodate the updating in Appendix C. For the fixed local regions case, we suggest that the number of local regions $K$ should be selected such that the average number of design points in each local region is sufficient to fit a GP model. A common recommendation is to fit a GP model with $10d$ design points in a $d$-dimensional space \citep{jones1998efficient,jalali2017comparison}. In sequential design problems, the number of initial fitting points can be smaller than $10d$ as more design points will be selected in the following iterations to improve the model fitting. For example, \cite{ankenman2010stochastic} used $4d$ initial design points in a one-dimensional queueing system. In CGLO, we use $4d$ initial design points in each local region on average. Therefore, $K$ can be decided as follows. If the total number of initial design points ($n_0$) is fixed, set $K=\lfloor n_0/4d \rfloor$; otherwise, the recommendations to adopt $K=10$ in \cite{meng2015additive} can be applied.  When $d$ increases, $K$ can be selected to increase with $d$, but it should be cautioned that this will increase the budget expensed for the initial design and this should be balanced with the remaining budget for the search. 

$\bf r_{min}$: This is a common parameter needed to be set for stochastic BO algorithms. Following the recommendation of \cite{Pedrielli2018etsso}, this can be decided when performing the cross-validation test with a small $r_{min}$ value and then increase it step by step until the cross-validation test is passed.

\subsection{Global Search Step}
This step is to quickly locate a promising region to concentrate the local search. The `promising regions' are regions that have lower global trends or have not been sufficiently searched (regions with the potential to find better solutions than those searched so far). Some popular acquisition functions are designed to select points in such regions by assigning larger acquisition function values to points with lower prediction mean or larger predictive variance (less explored). However, as the global model is based on the inducing points, the predictive variance $\hat{s}_g^2(\mathbf{x})$ \eqref{gmse} essentially reflects the spread of the inducing points instead of the actual design points. In particular, fewer inducing points from a region may not necessarily result from fewer design points. Instead, many design points with similar response values may be clustered together, and these points are summarized by a single inducing point. For such regions, there is little need to further explore. Hence, to balance the global search, we propose a global EI (gEI) criterion that incorporates the traditional EI function computed with the global model and a penalty term on the density of the design points nearby: 
\begin{equation}
gEI(\mathbf{x})=E\{\max(y_{gmin}-y_g(\mathbf{x}),0)\}\cdot\frac{1}{1+e^{n_a(\mathbf{x})/v-5}}\label{gei}
\end{equation}
where $y_{gmin}$ is the lowest global prediction at the inducing points, $n_a(\mathbf{x})$ is the number of neighbor design points of $\mathbf{x}$ and $v$ is a user-defined steepness parameter controlling how fast the penalty term decreases. In \eqref{gei}, $y_g(\mathbf{x})$ is normally distributed with variance $\hat{s}_g^2(\mathbf{x})$ and a mean of the following structure: $E[y_g(\mathbf{x})] = \hat{y}_g(\mathbf{x})$ if $\underline{M}\leq \hat{y}_g(\mathbf{x})\leq  \overline{M} $; $E[y_g(\mathbf{x})]=\underline{M}$ if $\hat{y}_g(\mathbf{x}) <\underline{M}$; $E[y_g(\mathbf{x})]=\overline{M}$ if $\hat{y}_g(\mathbf{x}) >\overline{M}$. Here, $\underline{M}$ and $\overline{M}$ are lower and upper bounds set for the predictive mean to avoid severe underestimation or overestimation caused by a limited number of design points or replications. These bounds do not need to be tight. If the response has natural bounds, $\underline{M}$ and $\overline{M}$ can be set as these values; Otherwise, they can be set as some reasonable constants (such a structure is similarly adopted in  \cite{sun2014}). The first term of the product in \eqref{gei} is the EI function representing how much we expect a new point to be better than the current best solution of the global model. The second term is a penalty designed diminishing with the number of design points around $\mathbf{x}$. The neighbors of $\mathbf{x}\in \mathbf{D_k}$ are defined as $\mathbb{B}(\mathbf{x})=\{\mathbf{x}'\in \mathbf{x}^k:||\mathbf{x}-\mathbf{x}'||<\kappa\}$, where $\kappa$ is the minimum distance between two inducing points used to define the neighborhood size, and $n_a(\mathbf{x})$ is the number of design points lying in $\mathbb{B}(\mathbf{x})$. 

This step optimizes $gEI$ over a finite global candidate points set $\mathbb{X}_{G}$, in contrast to the whole continuous domain $\mathbb{X}_{\Omega}$, which helps accelerate the optimization for this criterion. The rationale behind this is that in the Global Search Step, our purpose is to identify the best region and this does not require the accuracy of a search for a single best point. Therefore, we apply a finite candidate points set here, chosen with space-filling strategies from $\mathbb{X}_{\Omega}$ (to ensure that all local regions can potentially be selected, $\mathbb{X}_{G}$ should cover every region). From a practical viewpoint, this saves the computing budget for a more detailed search in the Local Search Step where the exact design points are selected. This idea of finite candidate points is also applied in other global optimization algorithms for searching criteria to efficiently manage computing budgets \citep{regis2007stochastic}.

\subsection{Local Search Step}
The local search is conducted in region $\mathbf{D}_{k,t}$, where $\mathbf{x}_g^0$, the best point in the global search step, belongs. This step searches more intensely within this promising region for a better solution. In order to achieve this, we apply the $mEI$ criterion \citep{quan2013simulation} to sequentially select the next design point,
\begin{equation}
mEI(\mathbf{x})=E(\max[y_{min}-z(\mathbf{x}),0])\label{mei}, \, \mathbf{x}\in \mathbf{D}_{k,t}
\end{equation}
where $y_{min}$ is the prediction at the best sampled points in $\mathbf{D}_{k,t}$, and $z(\mathbf{x})$ is normally distributed with variance given by spatial prediction uncertainty $\hat{s}^2_z=\tau^2_{k,t}-\mathbf{l'L^{-1}l}$ and a mean with the following structure: $E[z(\mathbf{x})] = \hat{y}_g(\mathbf{x})+\hat{y}_l(\mathbf{x})$, if $\underline{M}\leq \hat{y}_g(\mathbf{x})+\hat{y}_l(\mathbf{x})\leq  \overline{M} $; $E[z(\mathbf{x})]=\underline{M}$ if $\hat{y}_g(\mathbf{x})+\hat{y}_l(\mathbf{x}) <\underline{M}$; $E[z(\mathbf{x})]=\overline{M}$ if $\hat{y}_g(\mathbf{\mathbf{x}})+\hat{y}_l(\mathbf{x}) >\overline{M}$.
Here the overall AGLGP model is used as we require a more comprehensive and accurate search within this region for a better minimum point. We note that in $mEI$, the random noise is ignored in the predictive variance $\hat{s}^2_z$. The rationale behind this is to make $\hat{s}^2_z$ and $mEI$ values for already selected design points become zero such that they will not be reselected (as the Allocation Step will by design reduce the stochastic noise). This enables the search to focus on new points with low predicted responses and new points in the less explored areas of the local region. The point in $ \mathbf{D}_{k,t}$ that maximizes the $mEI$ function will be simulated with $r_{min}$ replications. The Local Search Step is sequentially repeated until the switching criterion is satisfied. 

Before introducing the switching criterion, we note that optimizing mEI (or other acquisition functions similarly) over continuous space requires an additional non-trivial optimization, which is a common issue for BO. The approaches to implement this include: (i). Use global optimization algorithms such as DIRECT \citep{jones1993lipschitzian}, and heuristics like GA. (ii).  In each step, randomly choose a discretization from the continuous space and choose the best one from this discretization (Jalali, 2017). In general, the first approach will require longer computational time, but has high accuracy. The second approach is faster and more economical from a practical viewpoint. In our experiments, we adopt the second approach and each time we optimize the mEI function in a local region, we use a fine discretization chosen with the Latin Hypercube strategy and maximize mEI on this discretization.

\paragraph{\textbf{Switching Criterion}}

In the local search step, the local model and $y_{min}$ are updated with the new observations sequentially. As we do not want to overly search a local region with the risk of getting trapped by local optima (i.e. spend excessive budget on negligible local improvements), we want CGLO to switch from local search to global search to locate potentially better regions when a local region is reasonably searched. Here, we adopt a `quality-based' switching criterion, which is commonly used for algorithm involving transitions between global and local phases or restart schemes \citep{regis2007stochastic,xu2010industrial,zabinsky2010stopping}.

As the best local region is decided by the $gEI$ value, a straightforward indicator to switch back to the global step to re-think the best region is when the $gEI$ values at points in the current best region become smaller than a threshold (indicating greater potentials in other regions). Recall that $gEI$ is defined to evaluate the global expected improvement of the region, which in other words evaluates the `quality' of solution improvements when searching in that region. Here, we propose a quality-based criterion to terminate the local step when $gEI(\mathbf{x}_g^0)$ is smaller than some threshold $G^*$:
\begin{equation}
gEI(\mathbf{x}_g^0)\leq G^*.
\end{equation}
Recall that the $gEI$ function \eqref{gei} has two components: the expected improvement and the density penalty. A small value for either one can result in a small $gEI$ value and the switching criterion may be satisfied. A small expected improvement (first term in \eqref{gei}) is likely to result from a low target value $y_{min}$ in \eqref{gei}, i.e., the current best has been largely improved in this local region. This indicates that a reasonably good or near-local-optimal solution has already been found. In addition, if the density penalty (second term in \eqref{gei}) is small, many points are observed in a small area, indicating the search is trapped in a small region, which is potentially a local optimal region. Hence, we would like to design the switching criterion such that when a reasonable solution or local optimal is found within a local region, opportunities are given to search in other potential regions with a switch back to the Global Search Step. 

In practice, we choose the threshold $G^*$ as $\max_{\mathbf{x}\in \mathbb{X}_{G} \cap \{\mathbb{X}_{\Omega}\backslash\mathbf{D}_{k,t}\}} gEI(\mathbf{x})$, which is the maximal $gEI$ value of global candidate points from other regions. We argue that this is a reasonable choice of $G^*$ to meet our requirement for the quality-based criterion. At the beginning of the Local Search Step, $\mathbf{x}_g^0$ has the best $gEI$ value among points in $ \mathbb{X}_{G}$. As we search more points in the local region and better and better solutions are found, $gEI(\mathbf{x}_g^0)$ decreases and a reasonably good solution can be found as explained above. When $gEI(\mathbf{x}_g^0)$ becomes smaller than this $G^*$, this indicates that another local region with better `quality' solutions is available. This prompts CGLO to switch back to the global step. We also note that even if this `reasonably good' solution found is not a local optimal or there exist multiple local optima in the current region and only one is found before switching back, CGLO by design can return back to this region for an even more detailed search in the future. This helps to balance the local search expense when the budget is finite and constrained, and further ensures global convergence, in contrast to some one-shot algorithms with local convergence guarantee, which never revisit or consider other local regions that have not been selected yet \citep{xu2010industrial}.

The CGLO algorithm can also be designed to incorporate other types of switching criteria such as an `effort-based' criterion where a limit on the number of new design points for each local step is set. Once the effort is exhausted, CGLO switches back to the global step. In practice, this `effort-based' criterion may further prevent over-exploitation and can be used in combination with the `quality-based' criterion, such that CGLO switches back to the global step when either of these two criteria is met.

\subsection{Allocation Step}
\label{sec:alloc}
The Allocation Step is dedicated to distributing additional simulations among sampled points to improve the accuracy of the noisy simulation evaluations. Specifically, it distributes for the following two purposes. First, to improve the global optimal estimate and ensure algorithm convergence, a minimum number of replications are required for each design point in every iteration. Second, to improve the local model fit and the optimal estimate, the noise of the observations should be drived down with more budget. 

Denote $M_t(\mathbf{x})$ as the total number of replications received by design point $\mathbf{x}$ in the first $t$ iterations. To achieve the first purpose, we state the following assumption on the allocation rule:

\begin{assumption}
	Suppose the noise variance $\sigma_{\epsilon}^2(\mathbf{x})$ among the input space $\mathbb{X}_{\Omega}$ is bounded, i.e., $\max_{\mathbf{x}\in \mathbb{X}_{\Omega}} \sigma_{\epsilon}^2(\mathbf{x})<\infty$. There exists a sequence $\{\kappa_1,...,\kappa_k,...\}$ such that $\kappa_{k+1} \geq \kappa_k$, $\kappa_k \rightarrow \infty$ as $k\rightarrow \infty$ and that $\sum_{k=1}^{\infty}k\exp(-a \kappa_k) <\infty$, $\forall a>0$.  It follows that $M_t(\mathbf{x})\geq \kappa_{N_t}$ for every of the $N_t$ design points in iteration $t$.
\end{assumption}

This assumption essentially requires sufficient budget to be allocated to each design point (specifically, at least $\kappa_{N_t}$ to each design point) in the first $t$ iterations. This rule helps to ensure global convergence of the algorithm (as will be shown in Section 5). To satisfy this assumption, at the beginning of the Allocation step, we check $M_t(\mathbf{x})$ for every design point and assign additional $\kappa_{N_t}-M_t(\mathbf{x})$ replications to the points where $M_t(\mathbf{x})<\kappa_{N_t}$. We refer to the budget consumed here for the first purpose in the Allocation Step as $B_{t,a}^1:= \sum_{i=1}^{N_t} \max\{0,\kappa_{N_t}-M_t(\mathbf{x})\} $. As the first purpose is to improve global estimates and convergence, the focus is taken on the design points in the entire design space, i.e., every design point in $\mathbb{X}_{\Omega}$ should satisfy the condition in Assumption 1. 

To address the second purpose, i.e., better fit of the local models, additional replications are distributed to design points in $\mathbf{D}_{k,t}$, as it is the most promising region in the current iteration. The maximum number of replications for this purpose is denoted as $B_{t,a}^2$, which is a user-defined input. OCBA provides an efficient way of allocating budget to identify the sampled point with the best response by balancing the response and noise level observed for each point. Denote the sample mean and variance at $\mathbf{x}_i$ as $\bar{y}_i$ and  $\widehat{\sigma}^2_{\epsilon}(\mathbf{x}_i)$, respectively. OCBA introduces the following allocation rule to maximize the Approximate Probability of Correct Selection asymptotically \citep{chen2000simulation}:
\begin{gather}\label{alloc}
\frac{N_{OCBA}(\mathbf{x}_i)}{N_{OCBA}(\mathbf{x}_j)}=\left(\frac{\widehat{\sigma}_{\epsilon}(\mathbf{x}_i)/\Delta_{b,i}}{\widehat{\sigma}_{\epsilon}(\mathbf{x}_j)/\Delta_{b,j}}\right)^2, i,j\in\{1,2,...N_t(\mathbf{D}_{k,t})\},\,  i\neq j\neq b\quad, \\
N_{OCBA}(\mathbf{x}_b)=\sigma_{\epsilon}(\mathbf{x}_b)\sqrt{\sum_{i=1,i\neq b}^n\left(\frac{N_{OCBA}(\mathbf{x}_i)}{\widehat{\sigma}_{\epsilon}(\mathbf{x}_i)}\right)^2}\notag,\\
\sum_{i\neq b}^{}N_{OCBA}(\mathbf{x}_i)+N_{OCBA}(\mathbf{x}_b)=B_{t,a}^2 \notag,
\end{gather}
where $\bar{y}_b$ is the best sample mean in $\mathbf{D}_{k,t}$, $N_{OCBA}(\mathbf{x}_i)$ is the number of simulations allocated to $\mathbf{x}_i$, $\Delta_{b,i}=\bar{y}_i-\bar{y}_b$ and $N_t(\mathbf{D}_{k,t})$ is the number of design points in $\mathbf{D}_{k,t}$. As mentioned before, $B_{t,a}^2$ is user-defined and since we focus here on illustrating the application of AGLGP model in optimization, we assign constant $B_{t,a}^2$ (the same value as $r_{min}$) over iterations for simplicity. More sophisticated adaptive budget allocation strategies can be applied to improve the algorithm \citep{chen2006efficient,peng2018ranking,Pedrielli2018etsso,peng2018efficient}.

In summary, in the Allocation Step, a check is first made on all existing design points to ensure at least $\kappa_{N_t}$ replications have been allocated to each point. If not, add replications to meet $\kappa_{N_t}$. Then, for design points in the current local region, allocate $B_{t,a}^2$ replications according to the OCBA allocation rule.

\section{Convergence of CGLO algorithm}
We next state the convergence results for CGLO. The proofs for all lemmas and the theorem are provided in the Appendices E, F and G. The proof is based on the following general assumptions.
\begin{assumption} 
	(a) The underlying objective function $f$ is bounded.

	(b) The hyperparameters ($\sigma^2,\tau_k^2$, $\pmb{\theta},\bl{\alpha}_k$) for the AGLGP model are bounded away from zero.
	
	(c) The noise variances $\sigma_{\epsilon}^2(\mathbf{x})$ are known.	
\end{assumption}

The EGO algorithm \citep{jones1998efficient} employs Assumptions 2(a) and 2(b) to ensure the bounded objective and the efficiency of the GP model. Known noises are assumed in Assumption 2(c) for analytical tractability. The convergence is only analyzed empirically \citep{kleijnen2012expected,Pedrielli2018etsso} if $\sigma_{\epsilon}^2(\mathbf{x})$ is unknown, to the best of our knowledge.

Recall that in the setting of CGLO, we propose predetermining or dynamically changing the number of local regions (see Appendix C). In both cases, the number of regions will be finite. With this in mind, we show CGLO algorithm convergence in three steps. We first show the design points selected are everywhere dense asymptotically. This is achieved by showing dense points in each local region (in Lemma 1) and showing each region can be selected as the promising region infinitely often (in Lemma 2). After that, with the guarantee in Assumption 1, we show not only the design points are dense, but the number of replications in each point is infinite. The convergence will then follow (in Theorem 1).

\begin{lemma}
	Under Assumption 2, the design points in local region $\mathbf{D}_{k}$ are dense if this region is given infinitely large budget.
\end{lemma}

\begin{lemma}
	Under Assumption 2, local region $\mathbf{D}_{k}$ will be selected infinitely often as the promising region. 
\end{lemma}

\begin{theorem}
	Under Assumptions 1 and 2, $\bar{y_t}^*\rightarrow f(\mathbf{x}^*)$ almost surely as $t\rightarrow \infty$, where $\bar{y_t}^*$ is the current best value and $\mathbf{x}^*=\arg\min_{\mathbf{x}\in \mathbb{X}_{\Omega}}f(\mathbf{x})$ is the true global optimal solution.
\end{theorem}

\section{Numerical Results}

In this section, we illustrate the proposed CGLO algorithm and compare it with other metamodel optimization algorithms in terms of both efficiency and computational time. For all the examples in this section, we apply $\kappa_k=0.1k$. We provide the selected algorithm parameters in Appendix H of the online supplementary material. 

\subsection{One-dimensional test function (illustration of algorithm)}

\begin{figure}[htbp]
	\centering 
	\includegraphics[scale=0.4]{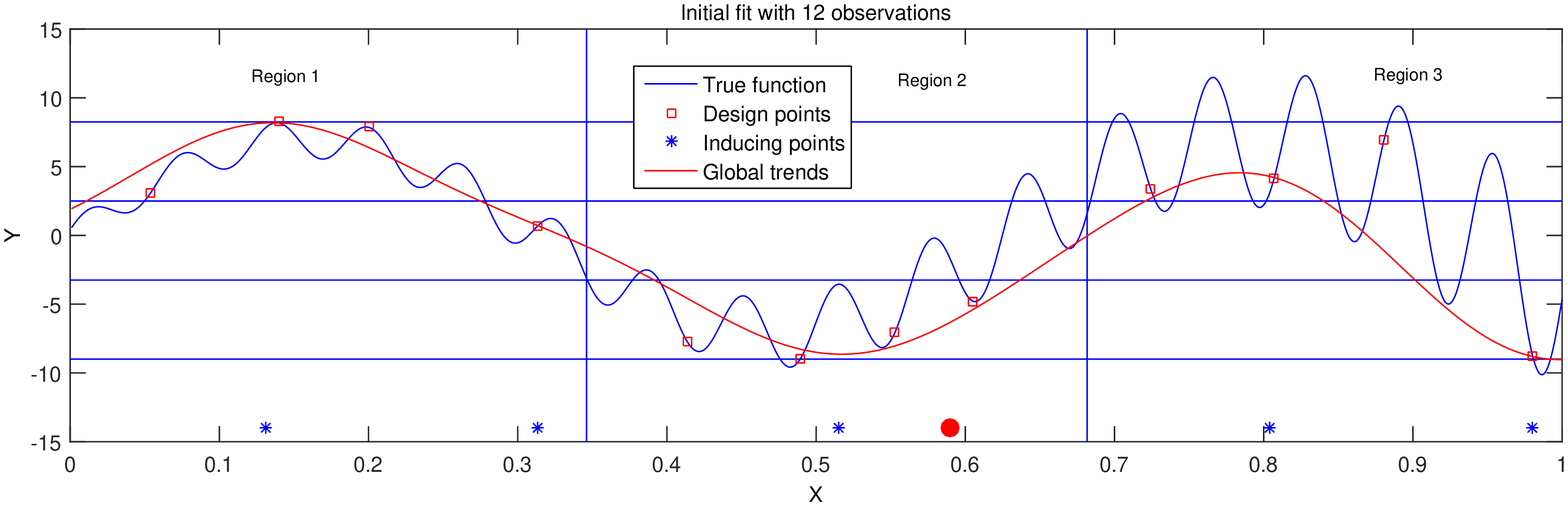}
	\caption{Initial fit}\label{initialfit}
\end{figure}

We applied the proposed algorithm to the illustrating example in Figure 1 with the noise variance $0.2+0.1\sin(10x)$ to further illustrate how the algorithm works. We initialize 12 design points with a Latin Hypercube design and thus set $K=3$ (see details about setting $K$ in Section 4), as seen in Figure \ref{initialfit}. As shown, the global optimal solution is -10.1316 at $x=0.9865$ and another sub-optimal solution is -9.5799 at $x=0.4826$. Hence, Regions 2 and 3 are more promising regions with lower response that should be focused on. Here, we set $r_{min}=B_{t,a}^2=20$. Two iterations of the algorithm were executed here for illustration.  

Figure \ref{initialfit} shows the estimated global model with the initial design points and we can see that it correctly identifies the overall global trend. As the global observations of inducing points are not observable, we only indicate the location of inducing points in Figure \ref{initialfit}. As shown, there are 3 local regions, each with two, one and two inducing points respectively. Based on the global model and inducing points, the Global Search Step selects Region 2 as the promising region (the location of $x_g^0$ is shown as the red $\bullet$ in Figure \ref{initialfit}). 
In the first round of local search, the point with the highest $mEI$ is evaluated with $r_{min}=20$ replications, and the clusters and the inducing points are re-generated in Region 2. With this additional point, however, the $gEI$ value at $x_g^0$ does not significantly decrease and the switching criterion is not met. Hence, the local search in Region 2 continues. After 4 new design points are added in Region 2, a few local optimal or near-optimal points are found (see Figure 3) and the switching criterion is satisfied. After that, CGLO stops the Local Search Step and executes the Allocation Step where 20 additional replications are allocated to the design points and then returns to the global search. 
\begin{figure}[htbp]
	\centering
	\includegraphics[scale=0.5]{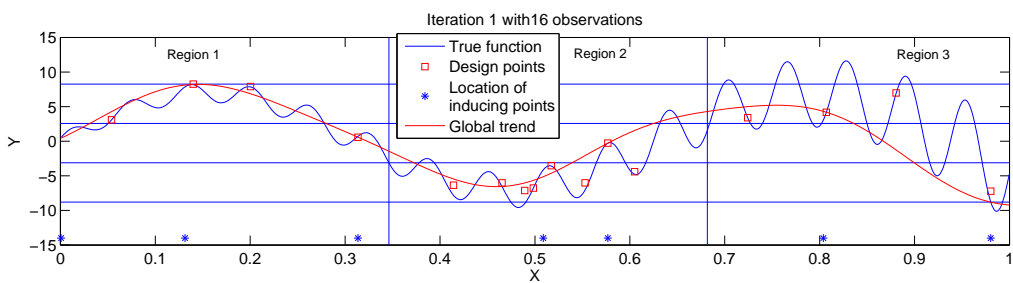}
	\caption{Iteration 1} \label{iter1}
\end{figure}

\begin{figure}[htbp]
	\centering	
	\includegraphics[scale=0.5]{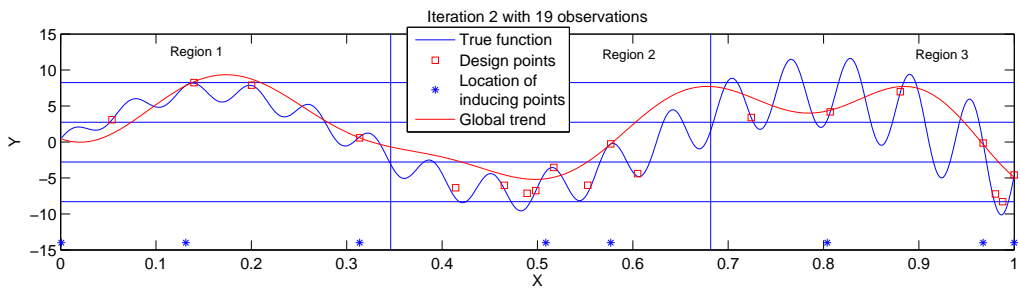}
	\caption{Iteration 2} \label{iter3}
\end{figure}

After the first iteration, a total of 16 observations are obtained as seen in Figure \ref{iter1}. The global model is updated, and as seen, the global model is now able to better capture the overall trend in Region 2, leaving Regions 1 and 3 less explored. As Region 3 has a lower trend between [0.9,1], the global search selects Region 3 as the promising region in the next iteration. CGLO then follows the same local search as iteration 1 and adds 3 more design points in Region 3 (Figure \ref{iter3}).

In the subsequent iterations, the search swings mainly between region 2 and region 3 and stops at region 3 with an estimated relative error (based on the estimated optimum and true optimum) of less than 1\% after 9 iterations. In this example, we see that the algorithm is able to identify promising regions efficiently with the global search and it is also able to search more intensely for the optimal solution in that region by the local search. 

\subsection{Comparative studies with other optimization algorithms}
In this section, we compare the CGLO algorithm with other BO algorithms. The test function is from \citet{sun2014}
\begin{equation}
\max_{0\leq x_1,x_2\leq 100}g(x_1,x_2)=10\cdot\frac{sin^6(0.05\pi x_1)}{2^{((x_1-90)/50)^2}}+10\cdot\frac{sin^6(0.05\pi x_2)}{2^{((x_2-90)/50)^2}}.\label{func}
\end{equation} 
The test function $g$ is multimodal. The global best is $g(90,90)=20$ and the second best is $g(70,90)=g(90,70)=18.95$. We can observe that the difference between the global best and the second best is quite small. To further increase the difficulty, we add a normal noise to this function: the mean is 0 and the variance is $\sigma^2_{\epsilon}(x_1,x_2)=3(1+x_1/100)^2(1+x_2/100)^2$.

\begin{figure}[htbp]
	\centering	
	\includegraphics[width=0.6\textwidth]{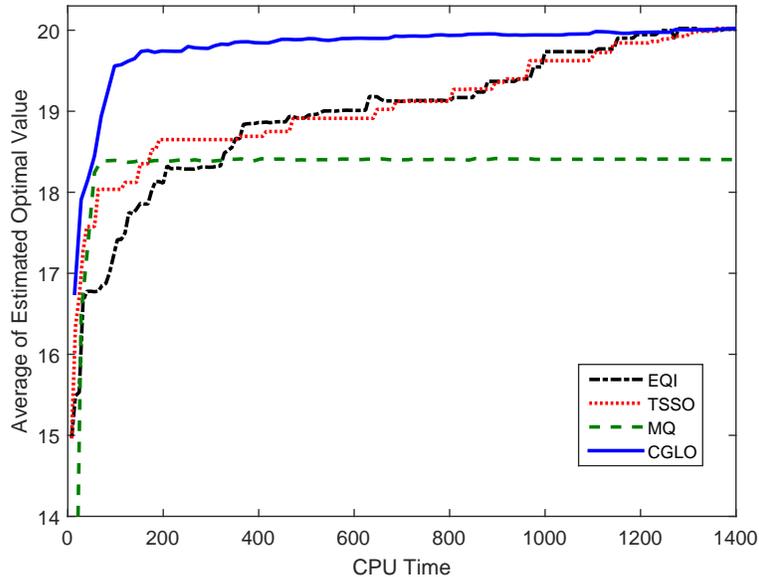}
	\caption{Estimated optimal value of TSSO, EQI and CGLO with CPU time of 1400s.}\label{800cpu}	
	
\end{figure}

To assess CGLO's effectiveness in addressing the challenges mentioned in Section 1, we compare it with three alternative BO algorithms, TSSO \citep{quan2013simulation}, Minimal Quantile criteria (MQ) \citep{jalali2017comparison}, and Expected Quantile Improvement (EQI) \citep{picheny2013quantile}, which are current BO methods for simulations with heteroscedastic noises. Among them, TSSO uses the OCBA allocation scheme and we select the online allocation scheme for EQI. We also note that knowledge Gradient (KG) \citep{frazier2009knowledge} is another competitive GP-based algorithm. However, KG requires much longer decision time to select the design points (about 50 to 100 times that of TSSO and CGLO) \citep{jalali2017comparison}. In this work, we are motivated by a real-time decision problem where the decision time is limited and in this comparative study, we use a relatively short fixed wall clock time of 1400 seconds for the comparison. Hence, KG is not included here. In each comparison, all algorithms start off with the same 40 LHS design points (and thus $K=5$ for CGLO) with 20 replications at each design point. We set $r_{min}=B_{t,a}^2=10$. Figure \ref{800cpu} shows the averaged optimal value estimated by each algorithm over 30 macroreplications.

From Figure \ref{800cpu}, we observe that CGLO converges much faster to the optimal value than TSSO, MQ, and EQI.   
We do however note that although CGLO numerically converges faster than TSSO and EQI, both TSSO and EQI both are able to catch up when a large amount of CPU time is expensed. MQ, on the other hand, does not converge and this is likely because it is stuck in one of the many local optima of this problem. 

To further compare these algorithms, Table \ref{1000average} presents their averaged performances (over 30 macroreplications) with different simulation replications budget (5,000 and 10,000 replications, respectively). Specifically, we consider the averaged distance between the observed optimal solution and the true optimal solution $|\Delta \mathbf{x}|$ and the average distance to the true optimal value $|\Delta y|$ for each of the algorithms. As seen, CGLO performs significantly better than TSSO, MQ, and EQI at $\alpha=0.05$ in terms of optimal value and optimal solutions when the budget is limited to only 5000 replications. With these limited replications, we see from the average of $|\Delta \mathbf{x}|$ that TSSO, MQ, and EQI can only find solutions in suboptimal regions while CGLO can provide solutions near the global optimal. This shows that the global and local natures of CGLO help it jump out of suboptimal regions to the global optimal region much faster than the traditional BO methods. When the simulation replication budget goes up to 10,000, there is no significant difference amongst CGLO, EQI, and TSSO. From this example, we see that the CGLO algorithm, even with the AGLGP model approximation, is able to achieve the level of accuracy similar to the TSSO and EQI in terms of optimal value and distance to the optimal solution. MQ, however, is still unable to locate the optimum, even with 10000 replications.

\begin{table}[htbp]\small
	\centering
	\caption{Average performance with 5000 and 10,000 simulation replications}\label{1000average}
	\begin{tabular}{p{1cm}p{0.5cm}|p{1.3cm}|p{1.3cm}|p{1.3cm}|p{1.3cm}|p{1.3cm}|p{1.3cm}|p{1.3cm}|p{1.3cm}}
		\hline
		{}&{}&\multicolumn{2}{c|}{CGLO}& \multicolumn{2}{c|}{TSSO}&\multicolumn{2}{c}{EQI}&\multicolumn{2}{c}{MQ}\\\cline{3-10}

		\multicolumn{2}{p{2cm}|}{No. of replications}&{average}& std &{average}& std&{average}& std &{average}& std \\\hline
		\multirow{2}{*}{5000}&$|\Delta \mathbf{x}|$&0.4821$^*$&0.2371&12.5764&13.3413&13.5894&14.2587&20.4929&14.9621\\
		&$|\Delta y|$&0.2298 $^*$&0.2005&0.8746&0.6880&1.1919&0.8992&1.6087&1.4465\\\hline
		\multirow{2}{*}{10,000}&$|\Delta \mathbf{x}|$&0.3369&0.1624&0.5166&0.2656&0.5158&0.2635&20.4839&14.9789\\
		&$|\Delta y|$&0.1991&0.2017&0.2106&0.1931&0.2099&0.1924&1.5928&1.4544\\\hline
	\end{tabular}
	\vspace{6pt}
	
	\raggedright{\small $*$ stands for better results in t-test at significance level $\alpha=0.05$}

\end{table}

\vspace{-0.3cm}
\subsection{A Navigational Safety Problem}

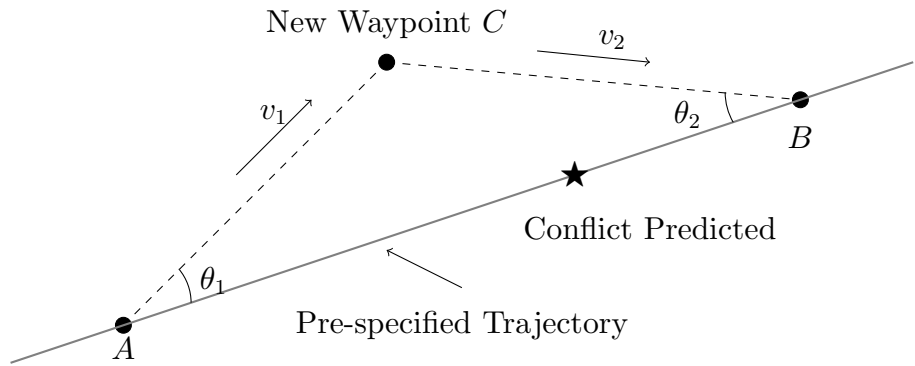
\begin{figure}[htbp]
	\centering
	\begin{tikzpicture}
	\draw[->] (0,1.5) -- (1,2.5);
	\draw[dashed] (-1.5,-0.5) -- (2,3);
	\filldraw[black] (-1.5,-0.5) circle (3pt);
	\draw[->] (4,3.15) -- (5.5,3);
	\draw[dashed] (2,3) -- (7.5,2.5);
	\filldraw[black] (7.5,2.5) circle (3pt);
	
	\draw[gray, thick] (-3,-1) -- (9,3);
	\filldraw[black] (2,3) circle (3pt) ;
	\node at (2,3.5){New Waypoint $C$};
	\node at (-0.3,0.1) {$\theta_1$};
	\node at (0.5,2.3) {$v_1$};
	\node at (6,2.3) {$\theta_2$};
	\node at (5,3.3) {$v_2$};
	\node at (-1.5,-0.8) {$A$};
	\node at (7.5,2) {$B$};
	\draw [->] (3,0) -- (2,0.5);
	
	\node at (3,-0.5) {Pre-specified Trajectory};
	\node at (4.5,1.5) {$\bigstar$};
	\node at (5.5,0.8) {Conflict Predicted};
	
	\draw (-0.6,-0.2)  arc[radius = 7mm, start angle= 0, end angle= 40];
	
	\draw (6.5,2.6)  arc[radius = 8mm, start angle= 180, end angle= 210];
	\end{tikzpicture}
	\caption{Definition of Trajectory}\label{traj}
\end{figure}

We next test our CGLO algorithm on a maritime safety problem. We apply it to the agent-based simulator developed for the Safe Sea Traffic Assistant ($S^2TA$) in \citep{ABM}, which looks 10 minutes ahead with an agent-based simulation model (ABM). For an own vessel travelling through a heavy traffic region, this ABM tries to predict any potential conflict with the other vessels based on the current and simulated environment. If a potential conflict of high risk is detected on a pre-specified trajectory of an own vessel from A to B (see Figure \ref{traj}), an optimizer is then called to find an alternative trajectory that minimizes the probability of conflict within this period.

\begin{figure}[htbp]
	\centering
	\begin{subfigure}[b]{0.5\textwidth}
		
		\begin{tikzpicture}
		\node at (0,0){\includegraphics[scale=0.5]{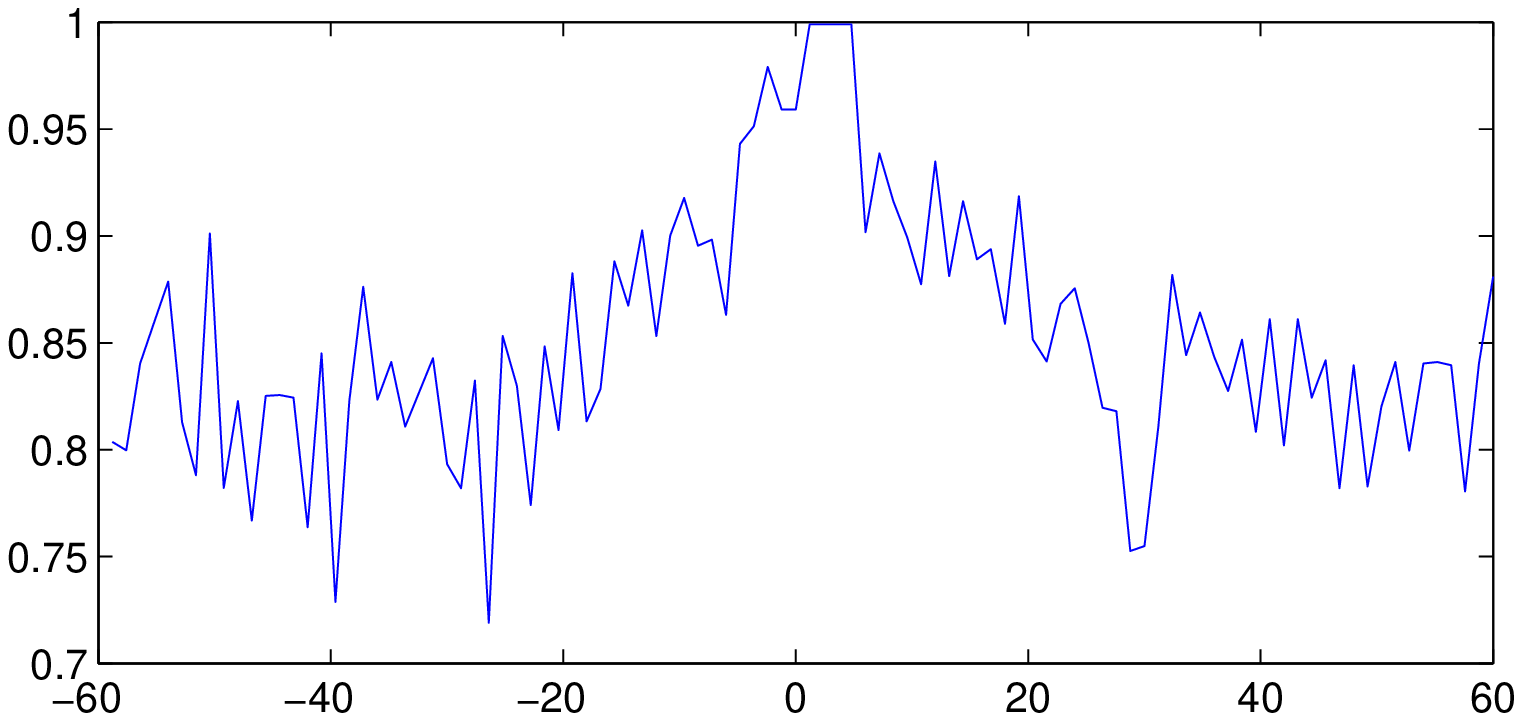}};
		\node at (0,-2) {$\theta_2$};
		\node at (-4.2,0) {$p$};
		\end{tikzpicture}
		\caption{}
	\end{subfigure}
	\begin{subfigure}[b]{0.49\textwidth}
		
		\begin{tikzpicture}
		\node at (0,0){\includegraphics[scale=0.35]{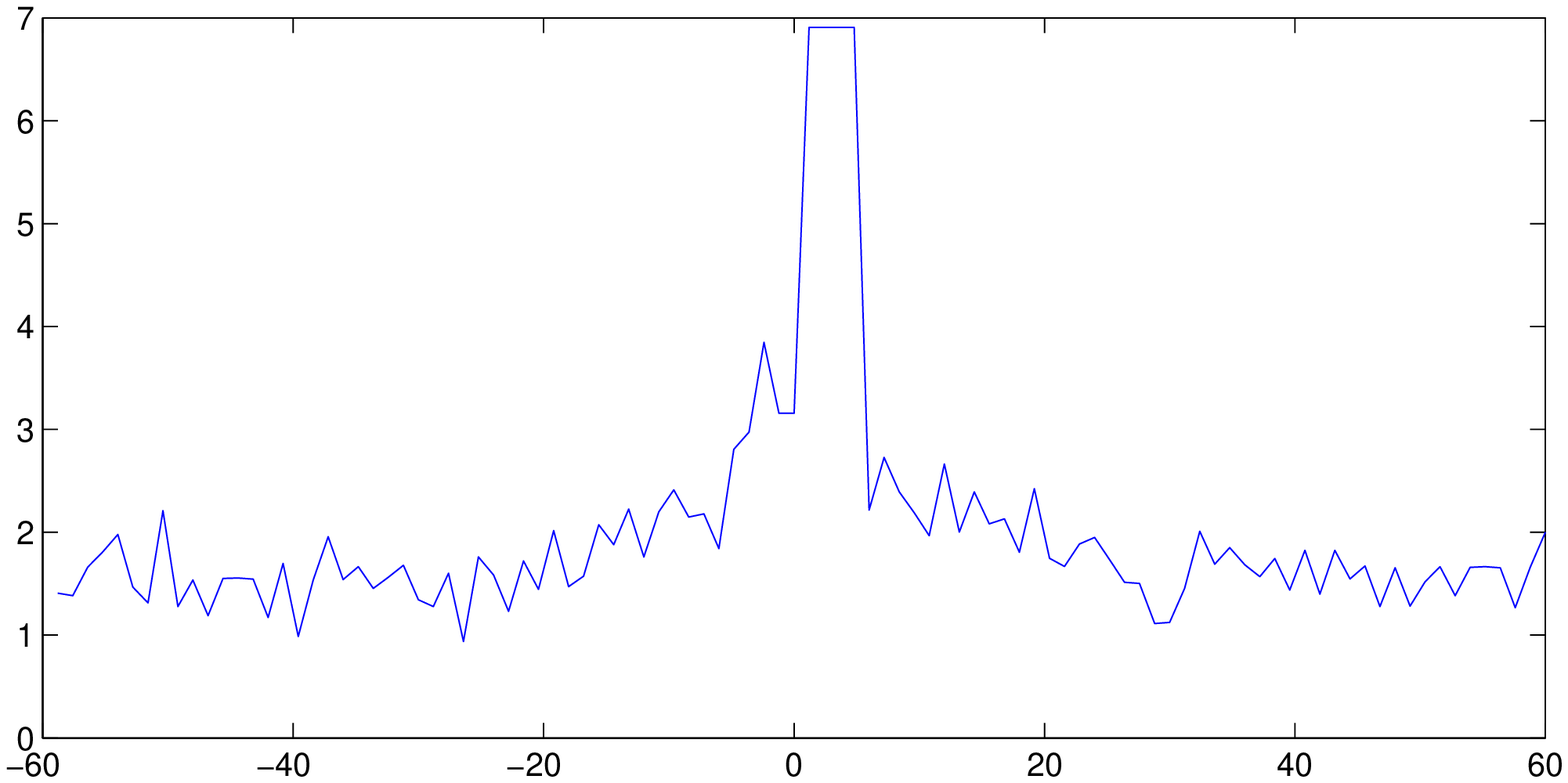}};
		\node at (0,-2) {$\theta_2$};
		\node at (-3.7,0) {$f$};
		
		\end{tikzpicture}
		\caption{}
	\end{subfigure}
	\caption{Probability of conflict in (a) and log transformation of probability in (b)}\label{pconf}
\end{figure}

Although $S^2TA$ looks at several objectives (including the probability of conflict, deviation from original trajectory) while searching for an optimal alternative route, we focus on the key safety objective of the probability of conflict in this example.  As this system focuses on heavy traffic regions, this response can be highly non stationary (as a small turn can result in a very different conflict environment). Figure \ref{pconf}(a) shows this response on one of the trajectory parameters for a vessel with 90 vessels in its proximity. The variance of this response also differs quite a bit in the negative and positive polar angles of the turn. 

In this system, an alternative trajectory is then defined by its deviation from the pre-specified trajectory. Specifically, the alternative trajectory is defined by a three-point trajectory of $A-C-B$, which is characterized by the deviation angles $\theta_1, \theta_2$ and the traveling speeds on each leg $v_1$ for $(A-C)$ and $v_2$ for $(C-B)$, see Figure \ref{traj}. For practical reasons, the range for $v_1$ and $v_2$ is between 1 and 15 [knots]. With positive and negative values on the angles indicating the deviations to the left or right from the original trajectory, $\theta_1, \theta_2$ are bounded between $-60^o$ to $60^o$ and constrained on the same polarity.

Due to the stochastic nature in the movement of surrounding vessels, to evaluate the probability of conflict of a candidate alternative trajectory, multiple replications of the simulator should be run to obtain the estimate. With the high nonstationarity of this response, the AGLGP model can be an appropriate metamodel for the response, and the CGLO algorithm an effective one to locate an optimal alternative trajectory.

As this objective ranges between [0,1], the following logistic transformation is applied, mapping the probability to the real line:
\begin{equation}
f(\mathbf{x}) = -ln\left(\frac{1}{p(\mathbf{x})}-1\right).
\end{equation}
With the transformation, the non stationarity of the response is still high (see Figure \ref{pconf}(b)).

As the $S^2TA$ is a real-time system, determining an alternative trajectory needs to be fast when a conflict is detected.  In this system, a 10-minute look-ahead period is applied, and a maximum of 5 minutes is allowed to obtain an optimal solution.  Fortunately, the simulator runs relatively fast, taking 0.006 seconds for one evaluation. In this example, relatively many evaluations can be afforded, but may still be insufficient to comprehensively search the entire four-dimensional space. We test CGLO with Random Search (RS) and TSSO in this study. Here, MQ is excluded due to its inferior performance (as seen in Section 6.2) and EQI is excluded due to its relatively long decision time. As observed both in our numerical runs and also in \cite{jalali2017comparison}, EQI takes about 3 times longer than TSSO to arrive at a decision and 6 times longer than CGLO, resulting in much fewer number of points that can simulated during the short time period given to make the decision. 

We start the system with a crowded scenario based on historical data gathered from the Strait of Singapore.  The system is run until a potential conflict is detected.  At this point, we run CGLO, RS, and TSSO to obtain the best optimal alternative trajectory. The time for the search is set at 5 minutes before termination, at which point, the final best solution is determined. To evaluate the solution obtained from each approach, the `optimal' solution was obtained from an extensive enumerative evaluation, which discretizes the solution space into $50 \times 50$ grids in both positive and negative polarity for $\theta_1$ and $\theta_2$, and the speeds $v_1$ and $v_2$ in $15\times 15$ grids and gives an estimation of minimum probability of $p^*=3.6165\times 10^{-4}$ at $\mathbf{x}^*=[-57.6000,-52.8000, 5.4783, 11.7682]$. 
The `optimal' solution was then compared with the solutions from the three approaches. Table \ref{conf_comp} presents numbers of design points and the distances between the `optimal' solution  $y_{true}$ and the observed best solution by the each approach $y_{approach}$.

\begin{table}[htbp]\small
	\centering
	\caption{Optimal Probability of conflict, the number of objective function evaluations, and optimal solution found by each optimization algorithm}\label{conf_comp}.
	\begin{tabular}{c|c|c}
		\hline
		Approach& Number of Evaluation Points& $\|y_{approach}-y_{true}\|$ 
		\\\hline
		{RS}&2000&0.0019
		\\
		{TSSO}&256&0.0117
		\\
		{CGLO}&611&$9.4876\times 10^{-4}$
		\\\hline
	\end{tabular}
\end{table}

As the simulation runs pretty fast, the random search can search at 2000 design points and identifies the optimal area that is close to the true optimal location. Although none of the approaches managed to locate the optimal solution in the limited time, as seen from Table \ref{conf_comp}, both RS and CGLO perform quite well. CGLO is able to locate a safer solution (with a 50\% lower probability of conflict) than RS.

Even though the TSSO works efficiently for expensive simulation optimization, a large part of the time is spent on model estimation for this problem, and only 7,325 replications are simulated on 256 design points. With limited design points, the TSSO only identifies a local optimal solution at the positive polarity. With the fast approximated AGLGP model and the combined search structure, the CGLO can approximately select 16,264 replications on 611 design points and find the optimal solution better with a more extensive search. 

A more extensive comparison is conducted from an additional 30 detected conflicts, and the averages obtained across the 30 conflict scenarios are shown in Table \ref{conf_ave}.
\begin{table}[htbp]\small
	\centering
	\caption{Average probability of conflict, number of function evaluations, distance to optimal solution for each optimization algorithm, and number of replications at each design points over 30 detected conflicts scenarios}\label{conf_ave}.
	\begin{tabular}{c|c|c|c}
		\hline
		Approach&Ave $\#$ of Evaluation Points& Ave $\|y_{approach}-y_{true}\|$ & Ave $\#$ of Replications at each point\\\hline
		{RS}&2000&$5.76\times 10^{-3}$ & 25\\
		{TSSO}&311&0.0189 & 29 \\
		{CGLO}&625&$1.58\times 10^{-3}$ &27\\\hline
	\end{tabular}
\end{table}
Here we see more conclusively that CGLO is able to find safer solutions with the global information of the response surface. TSSO, on the other hand, fails to work as efficiently for this problem and ends up with some local optima, similarly to the results in Section 6.2. These results thus show that there is a large potential for CGLO to be incorporated into the $S^2TA$ system to provide much faster evaluations than TSSO (and hence, providing a more extensive search), and safer solutions (in terms of probability of conflict) than RS.     

\section{Conclusion}
This paper proposes CGLO, a combined global and local search approach. The scheme first identifies promising regions with a global step and then focuses more deeply in these promising regions with the local step. We propose new EI-based search criteria to more effectively suit the global and Local Search Steps’ purposes, a built-in switching mechanism to switch between the steps to avoid overexploitation, and derive new allocation strategies to improve the metamodel fit and optimal solution estimation of the algorithm. Furthermore, we ensure the global convergence of the algorithm and study its performance on a test function and a practical example in maritime safety. With this proposed algorithm, we address some challenges faced by traditional BO approaches, especially when the baseline responses are highly multimodal. Firstly, the fast global approximation together with the local models of the proposed two-stage AGLGP is efficient and allows flexible model fit in different regions. Secondly, the global and local nature of the algorithm helps CGLO jump out of suboptimal regions towards global optimal faster. These improvements are very important for real-time decision processes, like in the navigational safety problem, where the response is highly multi-modal and response time is crucial. 

The current work can be extended in several directions. First, in CGLO, we currently apply a constant budget allocation scheme in the Allocation Step to distribute replications to points in the local regions. The algorithm can be improved with adaptive schemes that dynamically decides the allocation budget for better evaluation in each iteration. Second, in the Local Search Step, some alternative locally convergent approaches, such as pattern search, can be explored. As these methods are designed to perform well and fast in small regions, they can be more efficient than a GP metamodel approach in small local regions. Third, the local search in CGLO can be parallelized and adapted in the different local regions to further improve its efficiency.

\ACKNOWLEDGMENT{%
Wang is the corresponding author. A preliminary version of this paper was published in the Proceedings of the 2016 Winter Simulation Conference. Ng's work was partially supported by Singapore Ministry of Education (MOE) Academic Research Fund (AcRF) Tier 2 grant MOE2015-T2-2-148. Wang also acknowledges the support from Shenzhen Humanities and Social Sciences Key Research Bases, at Southern University of Science and Technology, College of Business (Shenzhen, China).
 
}

%

%
%


\bibliographystyle{informs2014} 
\bibliography{sample} 


\end{document}